\documentclass{article}

\usepackage{PRIMEarxiv}
\usepackage{natbib} 
\usepackage[utf8]{inputenc} % allow utf-8 input
\usepackage[T1]{fontenc}    % use 8-bit T1 fonts
\usepackage{hyperref}       % hyperlinks
\usepackage{url}            % simple URL typesetting
\usepackage{booktabs}       % professional-quality tables
\usepackage{amsfonts}       % blackboard math symbols
\usepackage{nicefrac}       % compact symbols for 1/2, etc.
\usepackage{microtype}      % microtypography
\usepackage{amsmath}        % for \text in math mode
\usepackage{lipsum}
\usepackage{float} % for [H] float placement
\usepackage{fancyhdr}       % header
\usepackage{graphicx}       % graphics
\usepackage{multirow}       % multirow
\graphicspath{{media/}}     % organize your images and other figures under media/ folder

\title{One Size Fits None: Modeling NYC Taxi Tips}

\author{
  Tomas Eglinskas \\
  The University of Texas at Austin \\
}

\begin{document}
\maketitle

\begin{abstract}
The rise of app-based ride-sharing has fundamentally changed tipping culture in New York City. We analyzed 280 million trips from 2024 to see if we could predict tips for traditional taxis versus high-volume for-hire services. By testing methods from linear regression to deep neural networks, we found two very different outcomes. Traditional taxis are highly predictable ($R^2 \approx 0.72$) due to the in-car payment screen. In contrast, app-based tipping is random and hard to model ($R^2 \approx 0.17$). In conclusion, we show that building one universal model is a mistake and, due to Simpson's paradox, a combined model looks accurate on average but fails to predict tips for individual taxi categories requiring specialized models.
\end{abstract}

\section{Introduction}

This paper was inspired by a simple question: if a new driver moved to New York City today, which taxi category should they drive to maximize their income? While the base rates usually stay fixed or regulated, the tip is always unknown before each trip's end. We wanted to determine if we could predict this volatility based on the taxi category.

As ride-sharing has become mainstream, customer perception of the tip has also changed. Since the introduction of yellow and green taxis, the in-car screen has been prompting passengers immediately at the end of the ride, while app-based services decouple the payment from the service entirely, usually taking the tip included in the overall service fee. We analyzed over 280 million trips from the 2024 datasets to see if these interface differences fundamentally change passenger tipping behavior. 

Testing methodologies range from regular linear regression to deep neural networks, eventually checking the hypothesis of whether we can make a universal model that could predict the tip amount despite the taxi category. 

\section{Research Background}
\label{sec:background}

Tipping is not just a machine learning problem; it is deeply rooted in psychology and economics. Past research shows that we can manipulate tip amounts simply by changing the interface, raising the default options by 5\%~\citep{haggag2014default} or altering how credit cards are presented~\citep{king2014explaining}. However, these prompts force passengers to actively think about the payment, which often creates a negative emotional response~\citep{goh2021pro}.

In this paper, we look specifically at 2024, as ride-sharing apps have adjusted their approach to tipping. Current guides suggesting that passengers are now expected to tip Uber drivers similarly to traditional taxis~\citep{WorldTodayJournal2024}. If these norms are indeed merging, we should be able to build a single, universal model that predicts tipping behavior across all existing taxi categories and successfully be used for predicting each taxi category.

\section{Data}
\label{sec:data}

For predictive analysis, we use the Taxi \& Limousine Commission (TLC) trip record data \citep{tlc_trip_data}. This data is publicly available because of a technical policy known as the "Open Data Law" \citep{nyc_open_data_law}, requiring all of New York's public data to be made available on a single web portal by the end of 2018, including the TLC. 

For each taxi category, the data is split into monthly datasets for the entire year and stored in the Apache Parquet format for compression. The TLC record data covers four different taxi categories:

\begin{itemize}
    \item \textbf{Yellow Taxis:} Taxicabs that have the right to pick up street-hailing and prearranged passengers anywhere in New York City. Each taxi must have a medallion affixed.
    \item \textbf{Green Taxis:} Taxicabs that can drop passengers off anywhere, but can only pick up passengers in outer boroughs (Queens, Brooklyn, the Bronx, Staten Island) and northern Manhattan. They were introduced by the city to improve access for other neighborhoods, as yellow taxis rarely visited and ignored other places beyond Manhattan and airports.
    \item \textbf{High-Volume For-Hire Services (HVFHS):} Vehicles dispatched by companies servicing more than 10,000 trips per day (e.g., Uber, Lyft). Today, this category represents the majority of taxi rides.
    \item \textbf{Standard For-Hire Vehicles (FHV):} Community car services and luxury limousines that operate solely via pre-arrangement and do not meet the high-volume threshold.
\end{itemize} 

Taxi trip records are collected differently by category. Yellow and green taxis use required in-vehicle hardware systems known as the \textbf{Taxicab Passenger Enhancement Program (TPEP)} \citep{tlc_rules_tpep} and the \textbf{Livery Passenger Enhancement Program (LPEP)} \citep{tlc_rules_lpep}. High-Volume For-Hire Services are mandated to electronically submit trip information through an API \citep{tlc_hvfhs}, and standard FHV rely on a periodic batch submission of dispatch logs \citep{tlc_fhv_submission}, which is usually executed manually or semi-manually. Unlike other categories, standard FHV data is primarily collected to track driver service hours and ensure compliance with fatigue regulations \citep{tlc_fhv_submission}. As a result, these categories of datasets do not have transactional information - fares and tips, as noted in the official documentation \citep{tlc_dd_fhv}, and we exclude this category from this predictive analysis.

This study analyses the remaining three service types (Yellow, Green, and HVFHS) using data from 2024, the most recent complete annual dataset. These datasets contain approximately 20--25 fields describing detailed trip characteristics, including timestamps, location IDs, and financial totals. The final combined dataset from the year category from 2024 contains \textbf{281,300,386} trips. Of these, \textbf{75,594,162} trips include a recorded tip, representing \textbf{26.87}\% of the total volume. 

\begin{figure}[h!]
    \centering
    \includegraphics[width=0.8\linewidth]{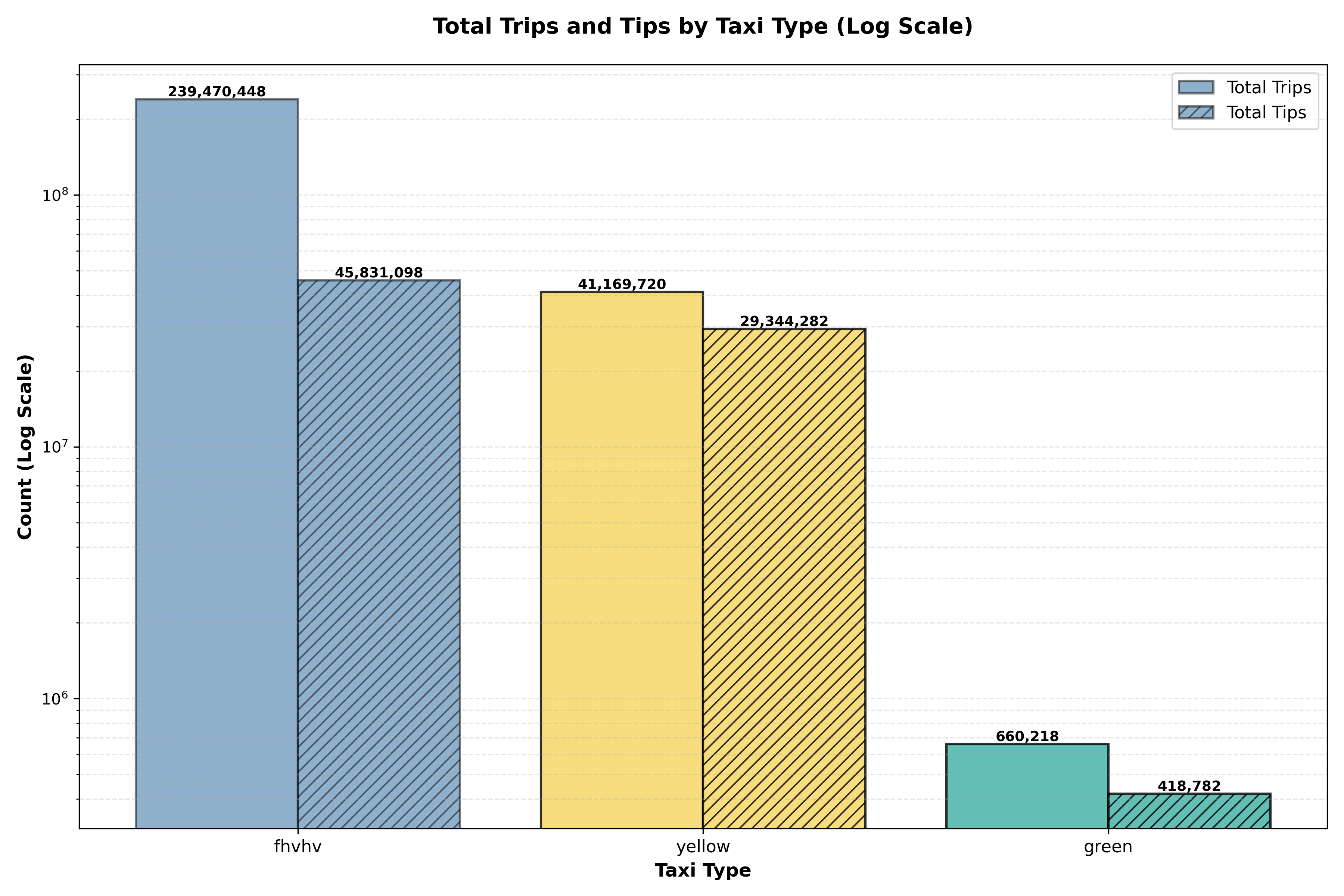}
    \caption{Total trips and with tips by category in 2024}
    \label{fig:taxi_types_trips}
\end{figure}

\subsection{Data Cleaning}
 
Before performing feature analysis and training models, we performed exploratory analysis on the three taxi type datasets. This step allowed us to better understand our data and identify necessary cleaning measures for mitigating the impact of extreme outliers. Since regression models are highly sensitive to anomalous values \citep{stevens1984outliers}, such as negative fares or physically impossible trip distances, defining strict inclusion thresholds was a prerequisite for model stability. Additionally, by cleaning values and defining a universal schema for the three taxi datasets, we created a universal dataset, which we use for our main analysis.
 
\subsubsection{Schema Merging}

As feature availability varies across services (e.g., green taxis lack \texttt{airport\_fee}, while HVFHS lacks \texttt{VendorID}, etc), we maintained the full, original feature sets for the individual service datasets without any modifications. To create the universal dataset, we merged all datasets for the three categories using only the nine shared features - a process known as data harmonization~\cite{cheng2024general}. This method allowed us to retain the majority of our data: the final combined set contains \textbf{276,039,083} trips, capturing \textbf{98.13}\% of the total \textbf{281,300,386} records.

\subsubsection{Outliers}
\label{subsubsection:outliers}

To mitigate the impact of extreme values during model training, we applied statistical filtering to remove outliers. First, we restrict trip distance ($d$) to the range $0 < d \le 26.20$ miles. The upper bound corresponds to the global 99th percentile, selected only to capture valid transit inside the city's territory and exclude records outside of it (e.g., trips recording unlikely distances such as \textbf{398,608} miles). 

Similarly, to address anomalies in fare data - such as single tips exceeding \textbf{\$470.03} - we cap tip amounts at the 99.99th percentile for each taxi category. Finally, we exclude the \texttt{cbd\_congestion\_fee} feature because it was introduced in 2025 \citep{tlc_trip_data} and all values for the year 2024 are \texttt{NULL}. These steps yield a dataset more representative of realistic New York City transit patterns, and for our analysis and models, we will have 4 datasets:

\begin{enumerate}
    \item \textbf{Yellow Taxis:} a dataset with \textbf{40,296,805} rows and 19 features.
    \item \textbf{Green Taxis:} a dataset with \textbf{624,406} rows and 19 features
    \item \textbf{High-Volume For-Hire Services (HVFHS):} a dataset with \textbf{236,706,978} rows and 24 features.
    \item \textbf{All Taxis:} a dataset with \textbf{273,218,046} rows and 9 features.
\end{enumerate}

\subsection{Feature Analysis}

\subsubsection{Distance Distribution}

Before training our models, we began by testing the fundamental hypothesis that trip distance is positively correlated with tip amount, a relationship observed in other taxi service analyses \cite{zhang2023prediction}. We analyzed the three distinct taxi datasets - yellow, green, and HVFHS individually, rather than using the combined set, to ensure we did not mask service specific behaviors. To avoid visual clutter and over-plotting common in large-scale datasets, we visualize these relationships using a random representative sample of $N=50,000$. The resulting correlations are shown below.
 
\begin{figure}[h!]
    \centering
    \includegraphics[width=1\linewidth]{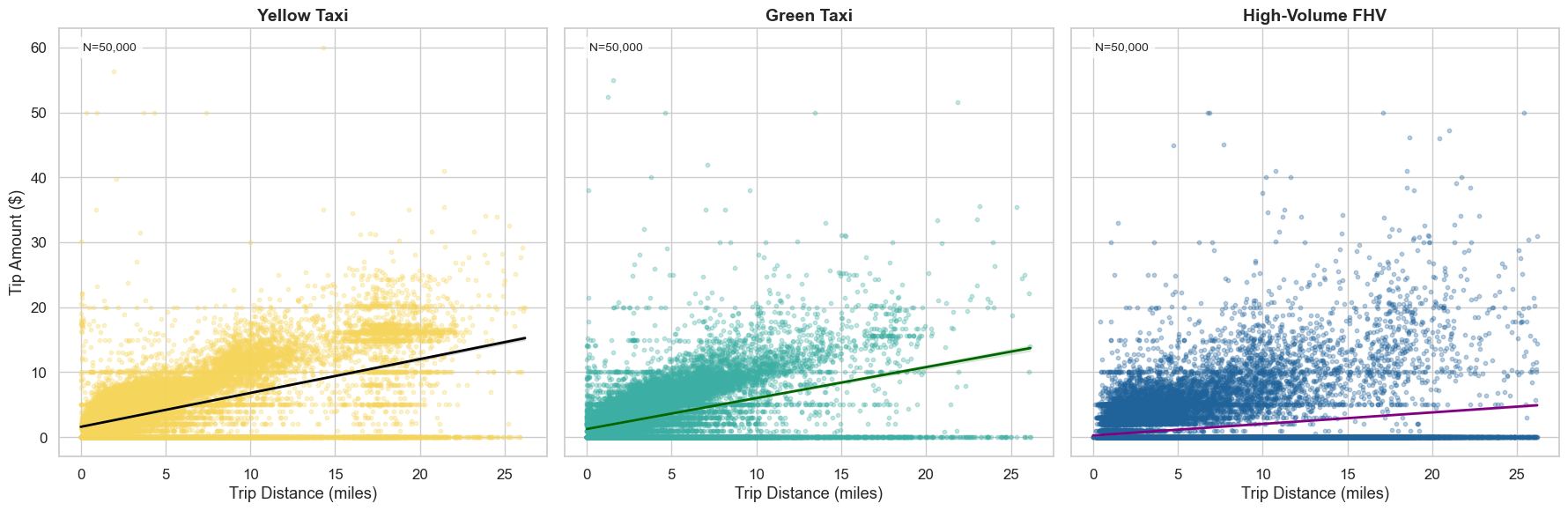}
    \caption{Tip and Distance Distribution across taxi types}
    \label{fig:taxi_types}
\end{figure}

To visualize the trend, we overlaid a simple linear regression fit on the scatter plots. As shown in Figure \ref{fig:taxi_types}, we see clear differences between the services. The yellow taxis plot shows an upwards slope, confirming a correlation - we could attribute this to percentage-based tipping (like 20\% or 25\%) found on the in-vehicle payment screen. Green taxis show a nearly identical pattern, which is expected as both taxi categories utilize the same metering hardware; the primary difference is simply the geographic zones where they are allowed to pick up passengers, as detailed in Section \ref{sec:data}.

In contrast, the HVFHS plot is significantly more scattered with a much flatter trend line. This indicates that trip distance is a poor indicator for tipping behavior compared to other taxi categories. Also highlighting a difference between taxis with in-vehicle versus in-app based tipping mechanisms and how they affect our psychology towards tipping, reducing the social pressure to tip proportionally~\citep{chandar2019drivers}~\citep{deguzman2019uber}~\citep{hanrahan2018roots}.

\subsubsection{Temporal Distribution}

To understand how time affects tipping, we used the same sample size of $N=50,000$ for each service. We created heatmaps to show two things: how many tips occur (Tip Count) and the typical tip amount (Median) for every hour of the week. A key finding is that for HVFHS, the median tip is \$0 regardless of the time. This means that most passengers simply do not tip. This confirms previous research suggesting that app interfaces reduce the social pressure to tip~\citep{chandar2019drivers}~\citep{hanrahan2018roots}, and it explains why we saw no clear pattern in the earlier distance plots (Figure~\ref{fig:taxi_types}).

\begin{figure}[h!]
    \centering
    \includegraphics[width=1\linewidth]{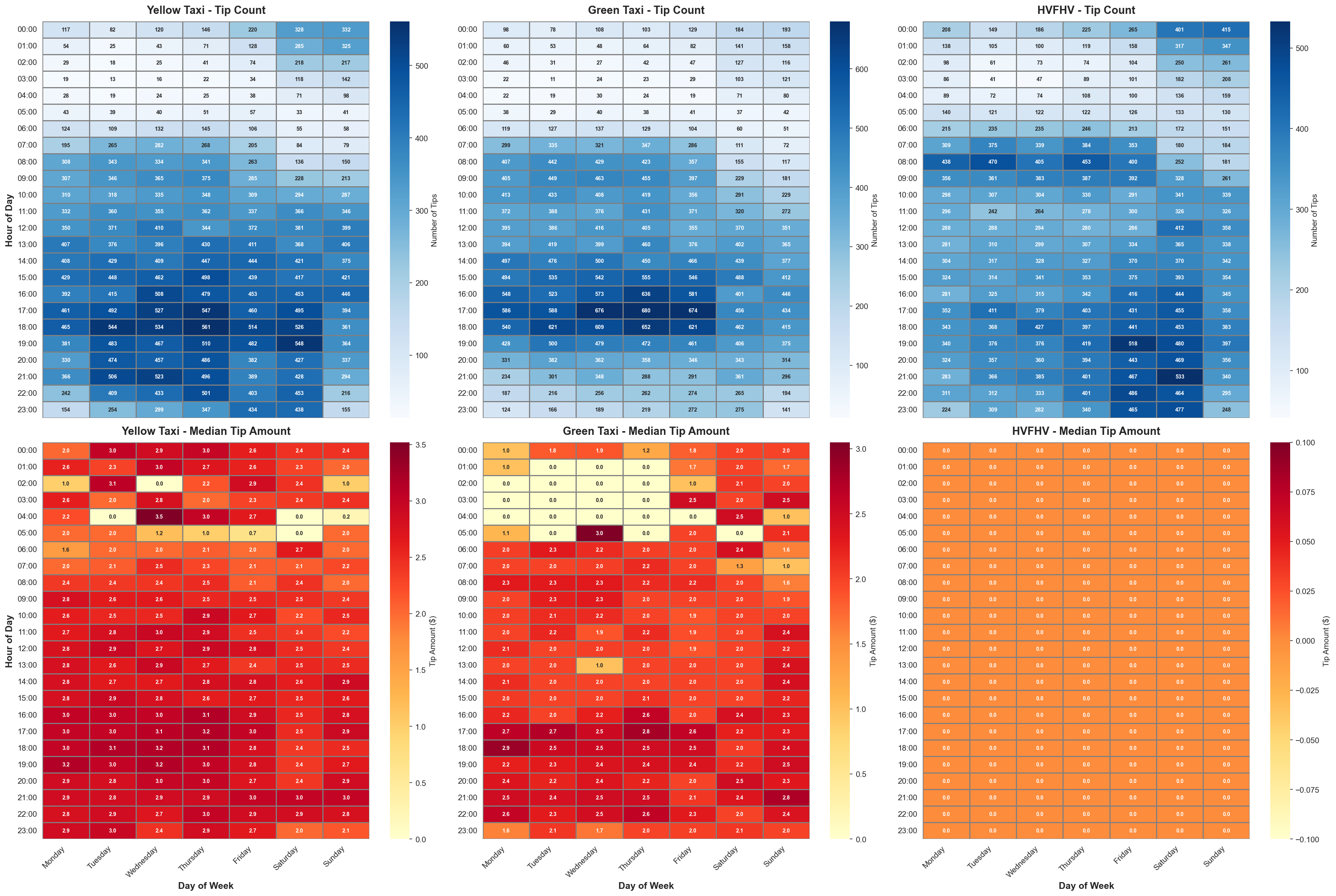}
    \caption{Heatmap of Tip Counts (Top) and Median Tip Amounts (Bottom)}
    \label{fig:tip_heatmap}
\end{figure}

The heatmaps (Figure~\ref{fig:tip_heatmap}) show us the best times to drive. For yellow taxis, the median tip is remarkably stable (around \$2.50--\$3.00) all day, with the busiest times being weekday evenings (18:00--22:00). Green taxis show the best time after work hours (16:00--18:00), which could indicate that for going out, most people use yellow taxis as they are allowed to pick up passengers in all Manhattan, where the main city center is. In contrast, for Uber/Lyft drivers, the time of day predicts how busy they will be (demand), but it does not predict if they will get a tip, since the baseline is almost always zero.

\subsection{Baseline Correlation Analysis}
\label{subsection:baseline}

To extend our analysis beyond two variable comparisons, we computed a Pearson correlation matrix \citep{hall1999correlation} which measures the correlation between all numerical variables at once. This allows us to spot redundant features and identify which variables have the strongest relationship with our target, \texttt{tip\_amount}.

However, as not all features are numerical values, we have dropped them from our initial matrix and will recreate these features in Section \ref{subsection:synthethic}. Consistent with our previous methodology, we utilized a random subsample of $N = 50,000$ records. 

\begin{figure}[h!]
    \centering
    \includegraphics[width=1\linewidth]{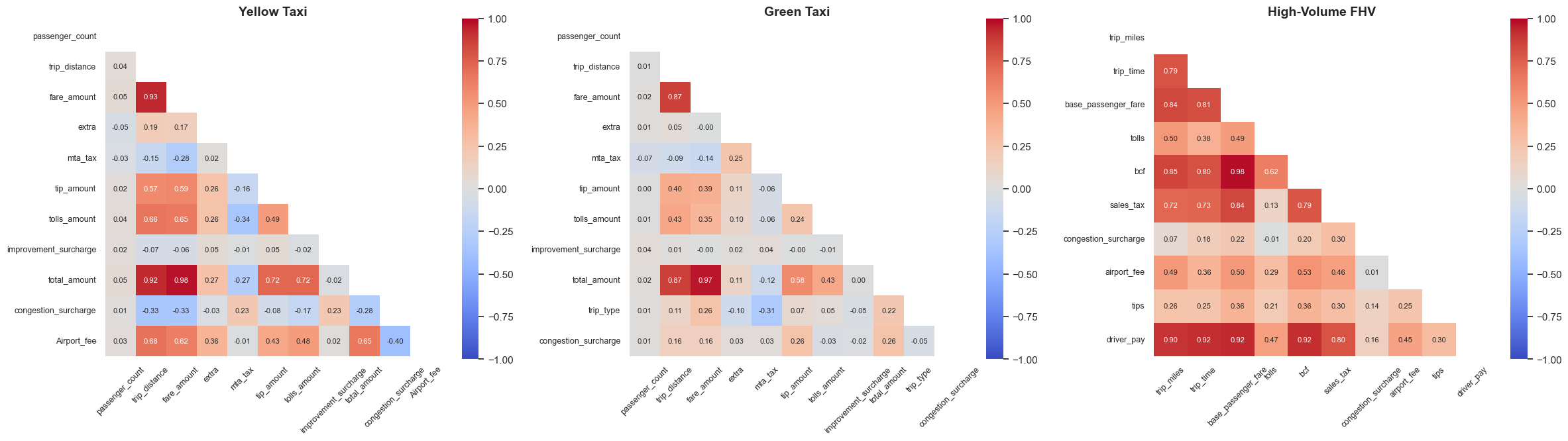}
    \caption{Correlation Matrix without Synthetic Features}
    \label{fig:cor_matrix_raw}
\end{figure}

We observed a correlation between \texttt{tip\_amount} and \texttt{total\_amount} across all categories. While statistically valid, this relationship is artificial: the \texttt{total\_amount} includes the tip itself. Leaving this in the dataset would cause target leakage, allowing the model to derive the tip perfectly just by doing math rather than learning passenger behavior - a critical error which is related to most common causes of overly optimistic predictive modeling results \citep{kapoor2023leakage}. Therefore, we dropped it based on the equation:

\begin{equation}
\text{Total Amount} = \text{Fare} + \text{Taxes} + \text{Surcharges} + \textbf{Tip Amount}
\end{equation}

By focusing on the non-leaking features, the analysis validates our hypothesis regarding service-specific tipping behaviors:

\begin{itemize}
    \item \textbf{Yellow Taxis} exhibit the strongest linear relationships. The tip amount correlates strongly with \texttt{fare\_amount} ($\rho = 0.59$) and \texttt{trip\_distance} ($\rho = 0.57$). This magnitude confirms that for traditional taxis, tipping is systematically linked to the bill size, consistent with the social norm of percentage-based gratuity.
    \item \textbf{Green Taxis} show a similar positive relationship, though slightly weaker. The correlation between tips and distance drops to $\rho = 0.40$, while the link to fare is $\rho = 0.39$. This likely reflects the different geographic service areas, as mentioned before, for only the outer boroughs.
    \item \textbf{HVFHS} presents a different view from the other taxi categories. While the driver's pay is almost perfectly linked to the miles driven ($\rho = 0.90$, HVFHS company pays more for more miles driven), the passenger's tip is not. The correlation between \texttt{tips} and \texttt{trip\_miles} falls to just $\rho = 0.26$. This supports the theory that the lack of a physical in-vehicle interface reduces the social pressure to tip proportionally to the fare.
\end{itemize}

These findings indicate that while linear models may predict yellow and green taxi tips effectively using fare and distance, predicting HVFHS tips will likely require non-linear approaches or additional behavioral features, as the linear signal is weak.

\subsection{Synthetic Features and Data Enrichment}
\label{subsection:synthethic}

A common technique for exploring and enhancing model performance is feature engineering from existing data or integrating external data source information. While the baseline correlation analysis in Section \ref{subsection:baseline} focused only on numerical variables, categorical variables like \texttt{DOLocationID} or \texttt{PULocationID} provide rich information that could be critical for our model's performance.

To create synthetic features for our categorical variables, we used one-hot encoding, which is a technique to convert one variable to multiple column features with values 1 indicating presence and 0 for absence~\cite{kanter2015deep}. Features like \texttt{DOLocationID} provide a numerical representation, but one \texttt{ID} is not greater in importance than the other by a higher value. Therefore, although this approach increases the dimensionality of our datasets, it allows us to investigate our data with more granularity before training and reuse the new features for training if they have importance~\cite{wang2021catboostmodelsyntheticfeatures}. We applied the following transformations to the categorical variables:

\begin{itemize}
    \item \textbf{Location Grouping:} \texttt{PULocationID} and \texttt{DOLocationID} were mapped to New York's five boroughs using the official \href{https://www.nyc.gov/site/tlc/about/tlc-trip-record-data.page}{TLC zone lookup table}. This created destination features \texttt{DO\_ZONE\_\{BOROUGH\}} and pickup features \texttt{PU\_ZONE\_\{BOROUGH\}}.
    \item \textbf{Airport Identification:} \texttt{RatecodeID} was remapped to a binary \texttt{is\_airport} feature.
    \item \textbf{Dispatch Status:} \texttt{trip\_type} was converted to \texttt{is\_dispatch} as in original datasets this feature is binary.
    \item \textbf{Payment Method:} \texttt{payment\_type} was expanded into \texttt{is\_cash}, \texttt{is\_card}, and \texttt{other}.
\end{itemize}

To expand our analysis with external features which could impact our model's accuracy~\citep{bernier2025}, we added a binary feature for national holidays (\texttt{is\_national\_holiday}) and historical weather data from \href{https://openweathermap.org/api}{OpenWeatherAPI} for 2024. The weather temperature was categorized into the following three features by the day's temperature:

\begin{itemize}
    \item \texttt{weather\_cold} ($< 10^\circ$C)
    \item \texttt{weather\_mild} ($10^\circ\text{--}20^\circ$C)
    \item \texttt{weather\_warm} ($> 20^\circ$C)
\end{itemize}

\begin{figure}[h!]
    \centering
    \includegraphics[width=1\linewidth]{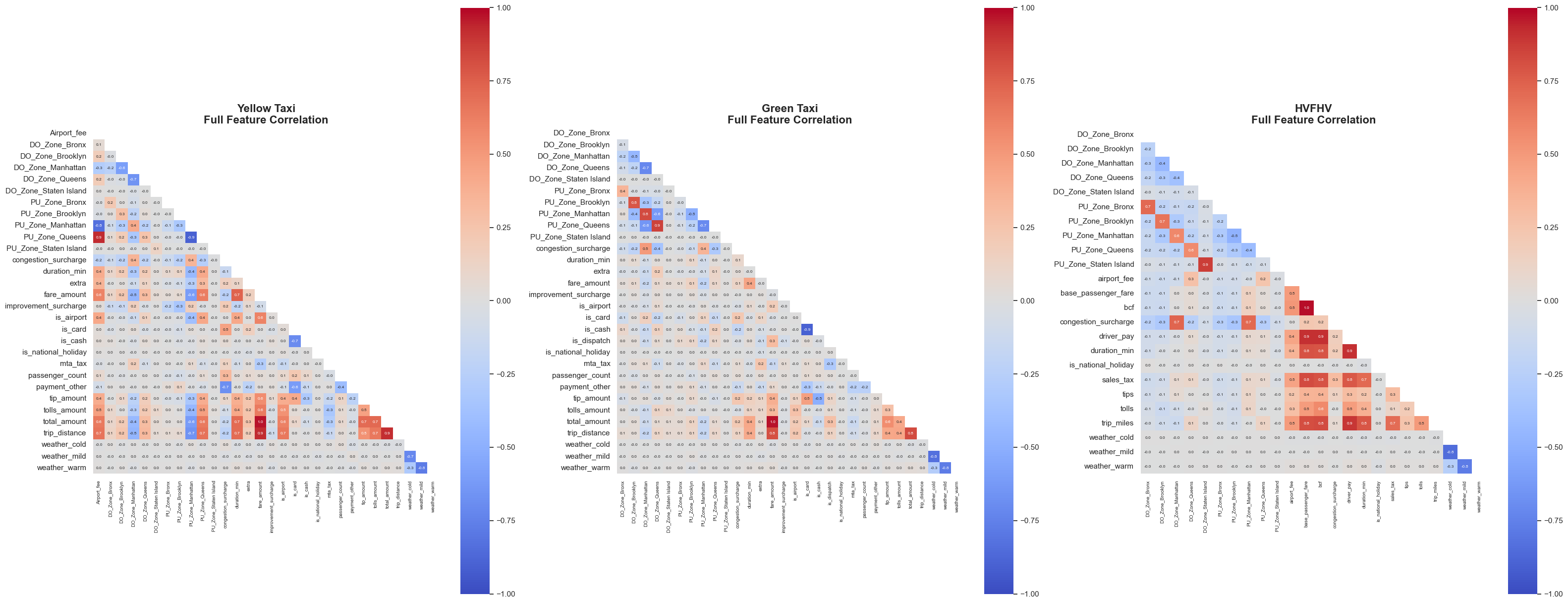}
    \caption{Correlation Matrix with Synthetic Features}
    \label{fig:cor_matrix_synthetic}
\end{figure}

The advanced analysis allows us to see how newly added features like pickup and destination location, payment methods, and weather affect tipping across different taxi categories (we will ignore our analysis on the leaked feature as mentioned in Section \ref{subsection:baseline}):

\begin{itemize}
    \item \textbf{Yellow Taxis} show that there is a strong dependency between the payment method and tipping. Paying by card is strongly linked to tipping, while paying with cash is linked to not tipping. This confirms our previously mentioned hypothesis that in-vehicle screens drive tipping - the screen forces you to choose a tip option, whereas cash passengers can easily skip it. Also, there are other correlations like \texttt{is\_airport}, which tells that people who drive to airports more often tip. 
    \item \textbf{Green Taxis} looks similar to the yellow taxis; however, with more features, we can more clearly see that the colors are more shallow. Inclining to believe that people from outer boroughs tip as yellow taxi passengers, but are less responsive to the in-vehicle prompts than the Manhattan-centric yellow taxis.
    \item \textbf{HVFHS} does not have obvious relationships between the newly introduced features, except for \texttt{is\_airport}, which suggests that, similarly to yellow and green taxis, people are psychologically inclined to tip for airport trips. As before, the matrix shows a near-perfect link between \texttt{driver\_pay} and \texttt{trip\_miles} ($\rho = 0.92$), proving that the app pays drivers accurately for the work they do. However, external factors like rain or cold weather show almost zero connection to \texttt{tips}. This suggests that while bad weather might make the ride more expensive (surge pricing), it does not actually make passengers tip more generously. 
\end{itemize}

Looking at the graph in Figure \ref{fig:cor_matrix_synthetic}, we can see that externally added features like weather and national holiday do not correlate with tipping behavior.

\section{Methods}

We begin by training a linear regression model for each of the three taxi categories and establishing a baseline. Eventually, we move towards other approaches and try to improve upon our performance calculations. After evaluating multiple different approaches, we pick the best approach and train our universal dataset on it, checking the performance on the taxi categories and our all-rounded dataset - all models use a split of 70:30 for training and evaluation and all metrics are reported on the evaluation sets.

All models are trained with datasets with previously created synthetic features in \ref{subsection:synthethic}, as this approach captures the most information about trips. Previously, we calculated that depending on the taxi category, the dataset size can vary from 600 thousand to 270 million trips. To reduce this amount for regression models, as usually it does not relate to an increase in accuracy~\citep{viering2022shapelearningcurvesreview}, we down-sampled the data by extracting a uniform random sample of 10\% from each month of the year, resulting in a balanced dataset of 1 million trips for yellow and HVFHS categories; for green taxi trips we retained the whole initial dataset.

For analysis metrics, we prioritize the mean absolute error (MAE) over the root mean squared error (RMSE). MAE offers a more robust representation of average model error in heavy-tailed distributions~\cite{willmott2005advantages}, unlike in Gaussian distributions~\citep{chai2014root}. In addition to the \texttt{tip\_amount}, we removed the \texttt{total\_amount} column to avoid feature leakage, as the total amount inherently includes the tip.

In all of the tables for model performance, we report:

\begin{itemize}
    \item \textbf{Mean Absolute Error (MAE)}: It tells us, on average, how much our prediction is off from the ground truth (in dollars). We prioritize it because it stays stable despite the extreme outliers.
    \item \textbf{Root Mean Squared Error (RMSE)}: We use this metric to check for model stability. Because RMSE squares the errors, it severely punishes big mistakes. If the RMSE is higher than the MAE, it signals that the model is making big prediction failures often.
    \item \textbf{Coefficient of Determination ($R^2$)}: This score tells us how much of the tipping behavior our model successfully explains.
\end{itemize}

\subsection{Algorithms}

\subsubsection{Linear Regression}

We established a baseline using linear regression to benchmark performance. As summarized in Table~\ref{tab:linear_regression_results}, this linear approach captures a moderate amount of variance for yellow taxis ($\mathbf{R^2} \approx \mathbf{0.610}$), yet it struggles significantly to generalize to the HVFHS dataset, resulting in a low $\mathbf{R^2}$ of \textbf{0.156}.

\begin{table}[h!]
    \centering
    \caption{Performance of linear regression}
    \label{tab:linear_regression_results}
    \begin{tabular}{lrrr}
        \toprule
        \textbf{Taxi Type} & \textbf{MAE} & \textbf{RMSE} & \textbf{$R^2$} \\
        \midrule
        Yellow   & 1.467 & 2.401 & 0.610 \\
        Green   & 1.379 & 2.242 & 0.484 \\
        HVFHS  & 1.631 & 2.824 & 0.156 \\
        \bottomrule
    \end{tabular}
\end{table}

An interesting insight is that the green taxi MAE ($\$1.38$) is smaller than the yellow taxi MAE ($\$1.47$) - meaning the model's predictions are closer to the true dollar amount for green taxis. However, the green taxi $\mathbf{R^2}$ is lower ($\mathbf{0.484}$ vs. $\mathbf{0.610}$). This occurs because green taxi tips have less natural variance, making it harder for the model to show a large improvement relative to simply predicting the average.

\begin{figure}[H]
    \centering
    \includegraphics[width=1\linewidth]{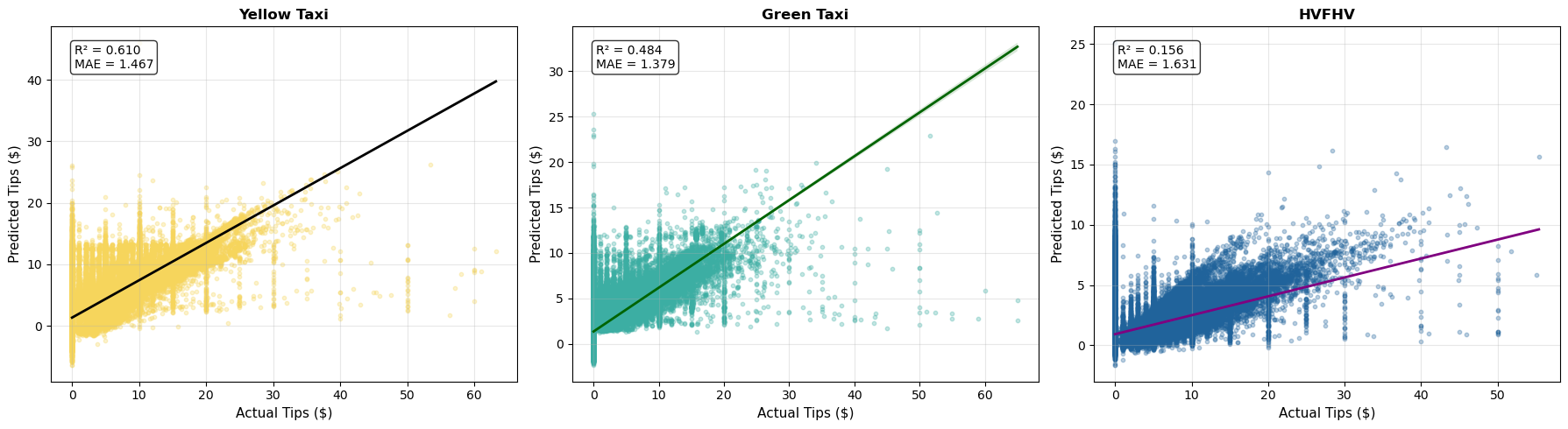}
    \caption{linear regression}
    \label{fig:linear_regression_plot}
\end{figure}

Figure \ref{fig:linear_regression_plot} visually demonstrates the limitations of the linear approach. While yellow and green taxis display a visible linear trend, the HVFHS data appear highly unstructured, as we have pointed out before. For HVFHS, the regression line does not follow any specific path; it merely runs through a disorganized scatter of points. This lack of structure confirms that simple linear models are insufficient for predicting ride-share tipping behavior.

\subsubsection{CatBoost Regressor}

Previously, for categorical features, we used explicit one-hot encoding, but we can use a CatBoost regressor ~\citep{prokhorenkova2018catboost}, which is a gradient boosting algorithm that natively handles categorical variables and was originally proposed by \cite{friedman2001greedy}. With gradient boosting, we can model non-linear relationships and explore more dependencies between different variables, making it potentially a better candidate for our tipping predictor.

As expected, gradient boosting substantially reduced our MAE (roughly 40\%) for all three taxi categories and improved RMSE for yellow and green taxis. We found that our model accounts for \textbf{68\%} of the variance in yellow taxi tips, achieving an MAE of $\$0.89$, and \textbf{56\%}  of the variance for green taxis, with a very similar MAE of $\$0.88$. 

\begin{table}[h!]
    \centering
    \caption{Performance of CatBoost regressor}
    \label{tab:catboost_results}
    \begin{tabular}{lrrr}
        \toprule
        \textbf{Taxi Type} & \textbf{MAE} & \textbf{RMSE} & \textbf{$R^2$} \\
        \midrule
        Yellow & 0.890 & 2.154 & 0.686 \\
        Green & 0.888 & 2.052 & 0.568 \\
        HVFHS & 1.093 & 3.263 & -0.126 \\
        \bottomrule
    \end{tabular}
\end{table}

However, the model produced a negative  $\mathbf{R^2}$ of $\mathbf{-0.126}$ for HVFHS, indicating that the available features provide no predictive value for our model predictions and are worse than just taking the mean tip value from all HVFHS trips.

\begin{figure}[h!]
    \centering
    \includegraphics[width=1\linewidth]{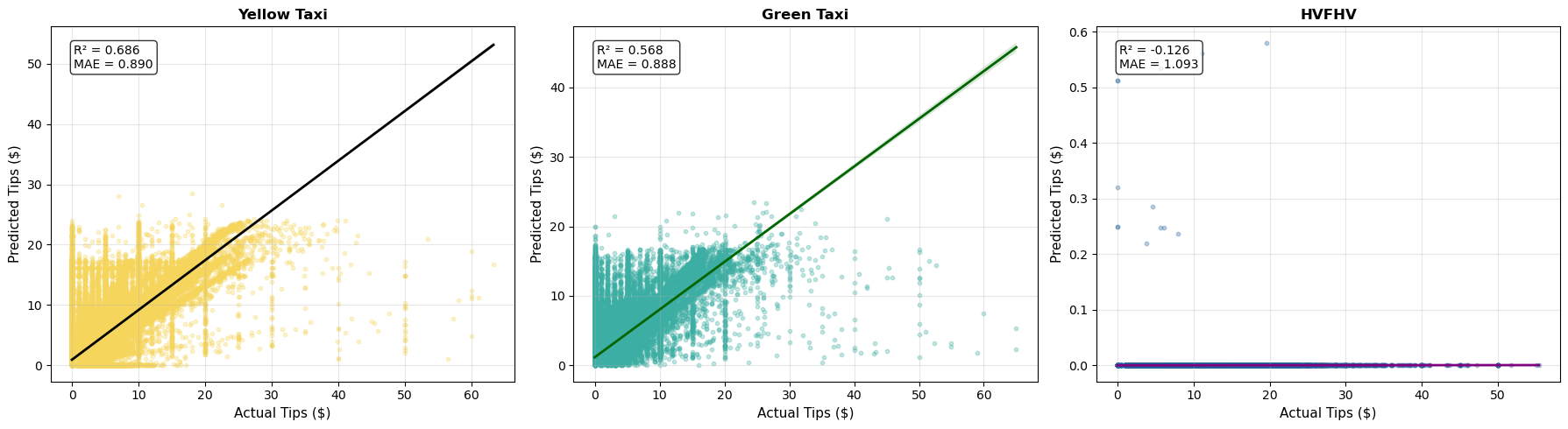}
    \caption{CatBoost Regression}
    \label{fig:catboost_plot}
\end{figure}

Figure~\ref{fig:catboost_plot} visually demonstrates these results - we have much tighter clustering around the diagonal axis for yellow and green taxis, particularly around the $\$15$ and $\$20$ dollars. And the HVFHS scatter plot indicates a failure to model the target, persisting as a vertical dispersion of errors with a negative score ($R^2 \approx -0.126$). The fact that a powerful non-linear ensemble model failed to impose structure on the HVFHS predictions strongly suggests that the poor performance is driven by a lack of predictive signal in the ride-share features rather than model incapacity.

\subsubsection{CatBoost with Tweedie Loss}
\label{subsubsection:tweedie}

Given that the tipping data is characterized by a high concentration of zero values and a heavy right tail, we further optimized the CatBoost model using the Tweedie loss function. Unlike standard RMSE or MAE optimization, the Tweedie distribution is specifically designed to model zero-inflated continuous data~\citep{zhou2019tweediegradientboostingextremely}.

\begin{table}[h!]
    \centering
    \caption{Performance of CatBoost with Tweedie Loss}
    \label{tab:tweedie_model_performance}
    \begin{tabular}{lrrr}
        \toprule
        \textbf{Taxi Type} & \textbf{MAE} & \textbf{RMSE} & \textbf{$R^2$} \\
        \midrule
        Yellow  & 1.049 & 2.046 & 0.717 \\
        Green   & 0.989 & 1.965 & 0.604 \\
        HVFHS   & 1.606 & 2.808 & 0.166 \\
        \bottomrule
    \end{tabular}
\end{table}

The use of the Tweedie loss function addressed the zero-inflated and heavy-tailed nature of the tipping data. This optimization resulted in a uniform improvement in \textbf{($\mathbf{R^2}$)} across all taxi types (Table \ref{tab:tweedie_model_performance}). While the MAE increased slightly across all categories, this was accompanied by a decrease in the RMSE for all categories. This specific tradeoff indicates the new model is less influenced by outliers and more robust against large prediction errors, explaining an additional $\mathbf{3.1\%}$ of the variance for yellow taxis compared to the standard CatBoost regressor.

\begin{figure}[h!]
    \centering
    \includegraphics[width=1\linewidth]{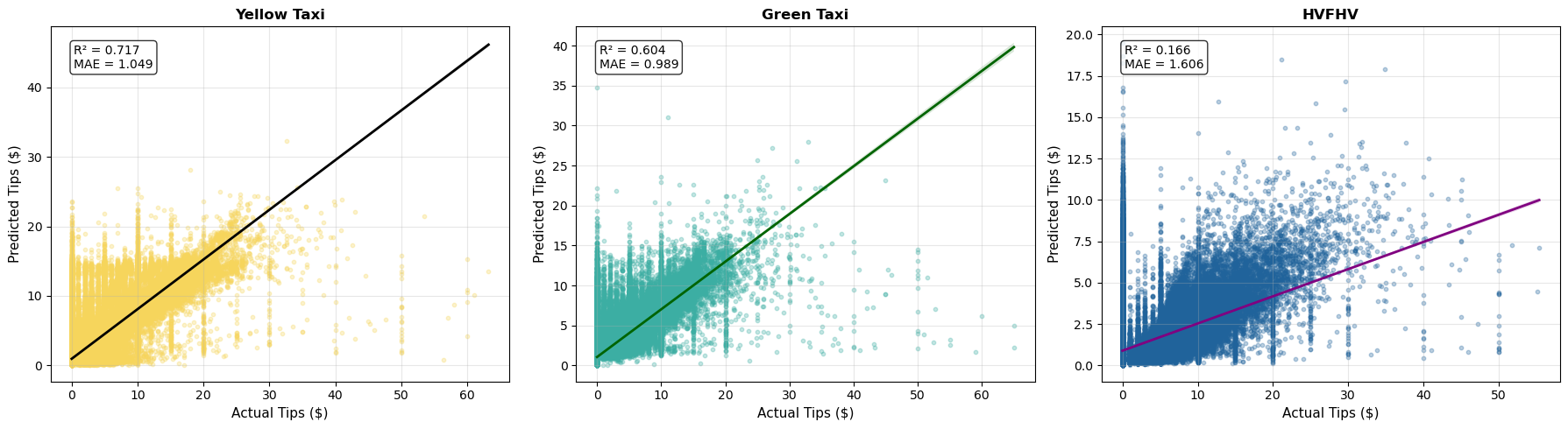}
    \caption{CatBoost with Tweedie Loss}
    \label{fig:tweedie_plot}
\end{figure}

\subsubsection{Deep Neural Network}

For interest, as our last model, we decided to create a deep feedforward network with 5 layers and a loss function for Mean Squared Error (MSE), training on 50 epochs. With this configuration, we adapted the neural network for a regression task as opposed to classification.

\begin{table}[h!]
    \centering
    \caption{Deep Neural Network Performance}
    \label{tab:final_model_summary}
    \begin{tabular}{lrrrr}
        \toprule
        \textbf{Taxi Type} & \textbf{MAE} & \textbf{RMSE} & \textbf{$R^2$} \\
        \midrule
        Yellow   & 1.049 & 2.044 & 0.717 \\
        Green   & 1.002 & 1.972 & 0.600  \\
        HVFHS      & 1.606 & 2.811 & 0.164   \\
        \bottomrule
    \end{tabular}
\end{table}

Interestingly, we have reached almost exact numbers as the yellow taxi and almost identical results for green taxi and HVFHS taxi categories as in the Tweedie loss model \ref{subsubsection:tweedie}, indicating that it effectively captured the underlying patterns. However, it did not improve further because the remaining variance is largely driven by noise and outliers in the tipping data, which are difficult to predict regardless of the model architecture or loss function used.

\begin{figure}[h!]
    \centering
    \includegraphics[width=0.8\linewidth]{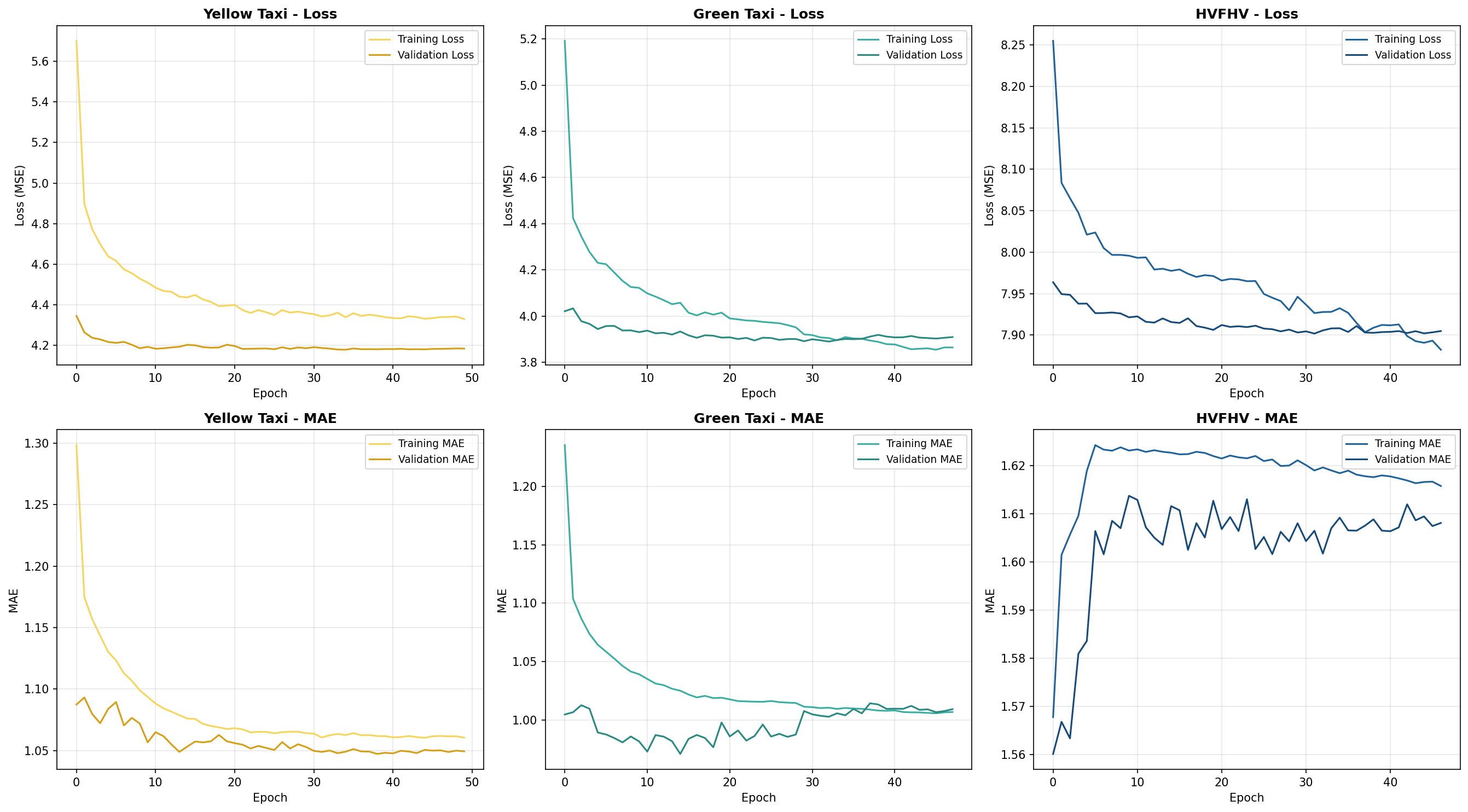}
    \caption{Deep Neural Network}
    \label{fig:dnn_training_history}
\end{figure}
 
The training history (Figure \ref{fig:dnn_training_history}) repeats these findings, showing rapid convergence primarily within the first 10 epochs. While the yellow and green taxi models exhibit stable generalization with validation loss closely tracking training loss, the HVFHS model displays volatility and a validation gap, reflecting the difficulty in modeling the high-variance for-hire data. Ultimately, the distinct plateauing across all metrics confirms that 50 epochs were sufficient to exhaust the model's learning capacity.

\subsection{Comparative Model Performance}
\label{comparative_all}

Table~\ref{tab:final_model_comparison} summarizes the different approaches and model results for each of the three taxi categories. The table highlights the distinction between traditional taxi categories and the High-Volume For-Hire Services.

\begin{table}[h!]
    \centering
    \caption{All Model Performance Summary}
    \label{tab:final_model_comparison}
    \begin{tabular}{llrrr}
        \toprule
        \textbf{Taxi Type} & \textbf{Method} & \textbf{MAE} & \textbf{RMSE} & \textbf{$R^2$} \\
        \midrule
        \multirow{4}*{Yellow}
            & Linear Regression & 1.467 & 2.401 & 0.610 \\
            & CatBoost Regressor & \textbf{0.890} & 2.154 & 0.686 \\
            & CatBoost with Tweedie & 1.049 & 2.046 & \textbf{0.717} \\
            & Deep Neural Network & 1.049 & \textbf{2.044} & \textbf{0.717} \\
        \midrule
        \multirow{4}*{Green}
            & Linear Regression & 1.379 & 2.242 & 0.484 \\
            & CatBoost Regressor & \textbf{0.888} & 2.052 & 0.568 \\
            & CatBoost with Tweedie & 0.989 & \textbf{1.965} & \textbf{0.604} \\
            & Deep Neural Network & 1.002 & 1.972 & 0.600 \\
        \midrule
        \multirow{4}*{HVFHS}
            & Linear Regression & 1.631 & 2.824 & 0.156 \\
            & CatBoost Regressor & \textbf{1.093} & 3.263 & -0.126 \\
            & CatBoost with Tweedie & 1.606 & \textbf{2.808} & \textbf{0.166} \\
            & Deep Neural Network & 1.606 & 2.811 & 0.164 \\
        \bottomrule
    \end{tabular}
\end{table}

For yellow and green taxis, non-linear methods outperformed the linear regression baseline. The CatBoost model with Tweedie loss and the deep neural network emerged as the superior architectures, increasing the explained variance ($R^2$) by over 10\% compared to the linear baseline (e.g., yellow taxi $R^2$ improved from 0.610 to 0.717). While the standard CatBoost regressor achieved the lowest MAE, the Tweedie and DNN models reduced it by approximately $\$0.40$ per trip compared to the baseline, maintaining a substantially higher RMSE and $R^2$, which is more important for real-life model use-cases.

In contrast, the HVFHS dataset showed resistance to model complexity. The standard CatBoost regressor failed to capture the signal, resulting in negative $R^2$ values (-0.126) due to the zero-inflated distribution. While the application of Tweedie loss and the DNN recovered the performance to match and slightly exceed the baseline ($R^2 \approx 0.166$ vs  $0.156$), the overall predictive power remained low. This stagnation confirms that the difficulty in predicting ride-share tips is likely a data-intrinsic issue-characterized by a weak signal-to-noise ratio-rather than a limitation of model capacity.

\section{Universal Model}
\label{sec:universal_model}

After analyzing the three different taxi categories and making models for their own categories, we combined all three category trips with a size of $N = 1,000,000$ and each containing 33\% of the whole dataset. Before even running the model on the evaluation split, we can hypothesize that the HVFHS dataset will be the biggest struggle, as before, and will decrease the overall performance for all existing taxi categories. 

We will use the CatBoost with Tweedie loss model ~\ref{subsubsection:tweedie} because of the best performing results ~\ref{comparative_all} and the speed of training this model versus a DNN, as in a production set, we would prefer this model for each of training and ease of deployment. 

\begin{table}[H]
    \centering
    \caption{Universal CatBoost Tweedie Performance}
    \label{tab:universal_tweedie_summary}
    \begin{tabular}{lrrr}
        \toprule
        \textbf{Taxi Type} & \textbf{MAE} & \textbf{RMSE} & \textbf{$R^2$} \\
        \midrule
        Universal & 1.232 & 2.322 & 0.567 \\
        \midrule
        Yellow  & 2.841 & 4.404 & -0.312 \\
        Green & 2.290 & 3.628 & -0.356 \\
        HVFHS & 1.167 & 3.223 & -0.095 \\
        \bottomrule
    \end{tabular}
\end{table}

\begin{figure}[h!]
    \centering
    \includegraphics[width=1\linewidth]{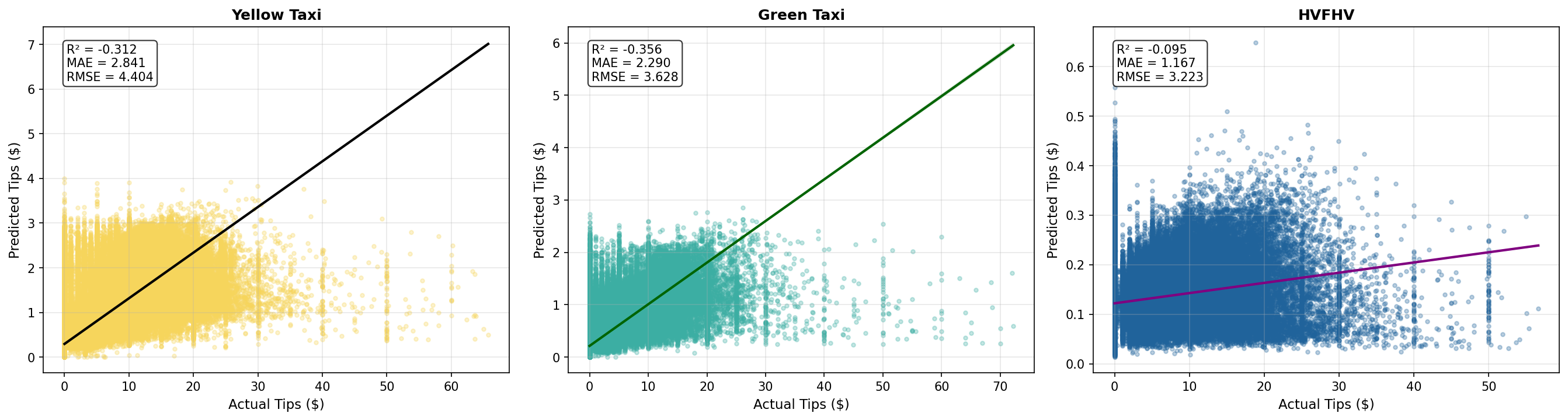}
    \caption{CatBoost with Tweedie Loss trained on universal dataset}
    \label{fig:universal_overall}
\end{figure}

Overall, the universal dataset performs quite well for the mixture of different taxi categories. However, we have to understand that two-thirds of the whole dataset contain yellow and green taxi trips, which are predictable. The limitations of this universal approach become clear when we break the performance down by vehicle type (Figure \ref{fig:universal_overall}). While the global metric ($R^2$ 0.567) is acceptable, it is heavily skewed by the volume of data rather than accuracy across the board.

This discrepancy - where the model appears predictive in aggregate but yields negative $R^2$ scores for every individual subgroup (Table \ref{tab:universal_tweedie_summary}) - is an example of Simpson’s paradox~\citep{sep-paradox-simpson} in regression analysis. The universal model has successfully learned to model the inter-group variance (e.g., distinguishing that yellow taxis generally have higher fares/tips than HVFHS rides) but has failed to capture the intra-group variance (the nuance of tipping behavior within a specific service type).

The model acts more as a broad classifier than a regressor, assigning an average for each taxi category rather than learning specific trends. This leads to performance worse than the simple mean within specific subgroups - a phenomenon known as \textit{negative transfer} \citep{pan2009survey}, where the dominant feature patterns from yellow taxis interfere with and degrade the learning of the distinct, high-variance HVFHS distribution. As shown by negative $R^2$ values (yellow $-0.312$, green $-0.356$) the error gap is substantial, the universal model's MAE for yellow taxis is \$2.841, whereas the specialized model achieves \$0.890. This discrepancy confirms that a unified architecture fails to capture the distinct economic dynamics of each service; separate models trained on specific service distributions are required for accurate prediction.

\section{Conclusion}
\label{sec:conclusion}

The evaluation of the universal model shows us the risks of mixing distinct data sources. On the surface, the universal model looks performant, however, a closer look reveals that a ``one-size-fits-all'' approach fails to capture the unique tipping habits found in different service types. We observed a statistical paradox - the global model shows a positive $R^2$, yet it produces negative $R^2$ scores for every single category. This proves that the economic and behavioral dynamics of yellow taxis, green taxis, and HVFHS are simply too different to be forced into a single predictive framework.

Our findings suggest the payment interface is a major factor in predictability. Yellow and green taxis use in-car terminals that push passengers with preset options, leading to strong correlations between the fare and the tip ($R^2 \approx 0.717$). In contrast, HVFHS payments happen in an app, often decoupled from the ride itself. The weak signal here ($R^2 \approx 0.166$) implies that for ride-shares, tipping is not driven by trip distance or cost, but rather by unobserved factors - such as passenger mood or post-ride interactions.

In conclusion, this study highlights that understanding the data distribution is often more critical than increasing model complexity. While standard linear regression failed to model the HVFHS category entirely, adding non-linearity via gradient boosting provided significant gains. Crucially, the CatBoost regressor with Tweedie loss emerged as the best approach. By explicitly handling the zero-inflated and heavy-tailed nature of the data, it matched the performance of a complex 5-layer deep neural network. The fact that these two very different architectures reached the same performance ceiling suggests we have hit the limit of the current feature set. Further improvements would require new data, not just deeper algorithms. This aligns with recent benchmarks showing that tree-based models often match or outperform deep learning architectures on tabular datasets \citep{grinsztajn2022why}. Ultimately, accurate forecasting in the New York City taxi market requires special models for each taxi category, rather than one general model.

%Bibliography
\bibliographystyle{unsrt}  

\bibliography{references}

\end{document}